\PassOptionsToPackage{linesnumbered,lined,boxed,noend}{algorithm2e}
\PassOptionsToPackage{nameinlink,capitalize}{cleveref}
\documentclass[pmlr,twoside,tablecaptionbottom,cleveref]{pgm-jmlr}
\usepackage{pgm-pmlr-dat}
\firstpageno{1}

\usepackage[utf8]{inputenc}
\usepackage[T1]{fontenc}

\usepackage{bm}
\usepackage{bbm}
\usepackage{mathtools}
\DeclareMathOperator*{\argsort}{arg\,sort}
\usepackage[outline]{contour}

\SetKwComment{Comment}{// }{}

\usepackage{xcolor}
\usepackage{graphicx}
\usepackage{float}
\usepackage{enumitem}
\usepackage{booktabs}
\usepackage{tikz}
\usetikzlibrary{arrows, arrows.meta}
\usetikzlibrary{decorations.text}
\usetikzlibrary{backgrounds}
\usetikzlibrary{shapes, shapes.geometric, positioning}
\usetikzlibrary{quotes}
\tikzset{
    >=latex,
    node distance=2.5cm,
    every node/.style={draw, circle},
    every path/.style={draw, thick, -},
    every edge quotes/.style ={
        draw = none,
        fill = white,
        execute at begin node = $,
        execute at end node   = $
    }
}

\usepackage{lipsum}

\usepackage{color}
\usepackage{soul}
\usepackage{mdframed}
\mdfsetup{backgroundcolor=yellow}

\usepackage{hyperref}

\title{Causal Discovery on Higher-Order Interactions}

\author{
    \Name{Alessio Zanga\nametag{\thanks{Corresponding author.}}} \Email{a.zanga3@campus.unimib.it}\\
    \addr Models and Algorithms for Data and Text Mining Laboratory (MADLab), \\
        Department of Informatics, Systems and Communication (DISCo), \\
        University of Milano - Bicocca, Milan, Italy \\
    \Name{Marco Scutari} \Email{scutari@bnlearn.com}\\
    \addr Istituto Dalle Molle di Studi sull’Intelligenza Artificiale (IDSIA), \\
        Lugano, Switzerland\\
    \Name{Fabio Stella} \Email{fabio.stella@unimib.it}\\
    \addr Models and Algorithms for Data and Text Mining Laboratory (MADLab), \\
        Department of Informatics, Systems and Communication (DISCo), \\
        University of Milano - Bicocca, Milan, Italy \\
}

\renewcommand{\hat}{\widehat}

\begin{document}

\maketitle

\begin{abstract}
Causal discovery combines data with knowledge provided by experts to learn the DAG representing the causal relationships between a given set of variables. When data are scarce, bagging is used to measure our confidence in an average DAG obtained by aggregating bootstrapped DAGs. However, the aggregation step has received little attention from the specialized literature: the average DAG is constructed using only the confidence in the individual edges of the bootstrapped DAGs, thus disregarding complex higher-order edge structures. In this paper, we introduce a novel theoretical framework based on higher-order structures and describe a new DAG aggregation algorithm. We perform a simulation study, discussing the advantages and limitations of the proposed approach. Our proposal is both computationally efficient and effective, outperforming state-of-the-art solutions, especially in low sample size regime and under high dimensionality settings.
\end{abstract}

\section{Introduction}

Causal graphs are one of the main drivers of modern causal inference
\citep{pearl2016causal, hernan2020whatif}. A causal graph is a directed acyclic
graph (DAG) used to describe the cause-effect relationships between a given set
of variables. A causal graph can be provided by domain experts, learnt from
available data, typically observational, or obtained by combining domain
experts' knowledge with the available data.

Causal discovery is the problem of recovering the causal graph by relying solely
on the available data or combining data with domain expert knowledge
\citep{glymour2019review, Zanga2022APractice}. In the last decade, causal
discovery has been applied to several scientific fields, including biology
\citep{sachs2005causal, Acerbi2014GeneDifferentiation,
Acerbi2016ContinuousHumans}, psychology \citep{miley2021schizophrenia}, medicine
\citep{Imoto2002BootstrapRegression, Zanga2022RiskApproach,
Zanga2023CausalStudy} and others \citep{Runge2019InferringSciences,
Hunermund2019CausalEconometrics, addo2021co2}. While the literature has
investigated how to learn a DAG from observational data
\citep{spirtes2000causation, spirtes2016causal, Huang2018GeneralizedDiscovery},
interventional data \citep{Triantafillou2015Constraint-basedSets,
kocaoglu2019advances} and incorporating prior knowledge
\citep{Constantinou2023TheLearning}, the causal discovery problem is far from
being solved. This is especially the case for real-world applications where data
have a small sample size \citep{Malinsky2018CausalGuide}, missing values
\citep{Scutari2020BayesianData, Tu2018CausalData, Liu2022GreedyValues} or hidden
variables \citep{bongers2021foundations}. In these settings, it is of paramount
importance to provide an estimation of the uncertainty of the recovered causal
graph. Approaches such as bagging provide a measure of \emph{confidence}
\citep{Friedman2003BeingNetworks, Eaton2007BayesianMCMC} in the inclusion of
individual edges into an {\em average DAG} obtained from a set of candidate
DAGs. This is an effective yet coarse summary of the complex cause-effect
relationships between variables due to its local nature: each edge involves only
two variables. To better mitigate the biases present in real-world observations
and preserve more complex patterns of causal relationships, \emph{we focus
instead on higher-order structures that involve multiple variables.} \\

\noindent The main contributions of this paper are:
\begin{itemize}
  \item the definition of \textit{higher-order structures}, that is, structures
    involving multiple variables, to represent complex edge patterns;
  \item the design of a novel model averaging algorithm where the aggregation
    step leverages the concept of higher-order structure;
  \item a simulation study showing the advantages and limitations of the
    proposed approach, especially in the small sample size regime and under high
    dimensionality settings.
\end{itemize}
The rest of the paper is organized as follows. \Cref{sec:preliminaries} provides
the necessary background. \Cref{sec:model_averaging} introduces our framework
for model averaging on higher-order structures and the corresponding aggregation
algorithms, which are evaluated in \Cref{sec:experiments} in a large simulation
study. Finally, \Cref{sec:conclusions} presents our conclusions and future
research directions.

\section{Preliminaries}
\label{sec:preliminaries}

\emph{Bayesian networks} \citep[BNs;][]{koller2009probabilistic} are a
foundational model to encode the interactions between random variables.

\begin{definition}[Bayesian Network]
  Let $\mathcal{G}$ be a DAG, let $\mathbf{X}$ be a vector of random variables
  and let $P(\mathbf{X})$ be a probability distribution with parameters
  $\Theta$. A Bayesian network $\mathcal{B} = (\mathcal{G}, \Theta)$ is a
  probabilistic graphical model where each variable in $\mathbf{X}$ is
  associated with a vertex of $\mathcal{G}$ and the global distribution
  $P(\mathbf{X})$ factorizes into local probability distributions according to
  $\mathcal{G}$:
  \begin{equation*}
    P(\mathbf{X}) = \prod_{X \in \mathbf{X}} P(X \, | \, \Pi_X),
  \end{equation*}
  with $\Pi_X$ the \emph{parents} set of $X$, the set of vertices with an edge
  into $X$.
\end{definition}

Since $\mathcal{G}$ is a DAG, each edge between two vertices $(X, Y)$ is a
directed edge denoted as $X \rightarrow Y$, or $\mathbf{e}_{XY}$ for short.
While BNs are probabilistic models, we redefine them as causal models as
follows.

\begin{definition}[Causal Edge]\label{asm:causal_edge}
  Let $\mathcal{G}$ be a DAG, and let $\mathbf{X}$ be a vector of random
  variables. The value assigned to each variable $X \in \mathbf{X}$ is
  completely determined by the function $f_X$ given its parents $\Pi_X$:
  \begin{equation*}
    X \coloneqq f_X(\Pi_X) \qquad \forall X \in \mathbf{X}.
  \end{equation*}
\end{definition}

This definition allows us to interpret the edges of $\mathcal{G}$ in a
non-ambiguous way: it enforces a recursive relationship over the structure of
$\mathcal{G}$, establishing a chain of functional dependencies. If $\mathcal{G}$
satisfies \Cref{asm:causal_edge} then it is called \emph{causal graph}
\citep{Pearl1995FromNetworks}.

\begin{definition}[Causal Network]
  A causal network is a BN where $\mathcal{G}$ is a causal graph.
\end{definition}

Following \cite{Scutari2013OnModelling}, given a BN model $\mathcal{B} =
(\mathcal{G}, \Theta)$, its posterior distribution given the data
$P(\mathcal{G}, \Theta \, | \, \mathcal{D})$ decomposes into  the product of
$P(\mathcal{G} \, | \, \mathcal{D})$, the probability distribution of the graph
$\mathcal{G}$ given the data $\mathcal{D}$, and $P(\Theta \, | \, \mathcal{G},
\mathcal{D})$, the probability distribution of the parameters $\Theta$ given the
graph $\mathcal{G}$ and the data $\mathcal{D}$. Therefore, the problem of
learning a BN model $\mathcal{B}$ from the available data $\mathcal{D}$
decomposes as follows:
\begin{equation*}
    \underbrace{P(\mathcal{G}, \Theta \, | \, \mathcal{D})}_\text{Learning}
    \quad = \,
    \underbrace{P(\mathcal{G} \, | \, \mathcal{D})}_\text{Structure Learning}
    \, \cdot \quad
    \underbrace{P(\Theta \, | \, \mathcal{G}, \mathcal{D}).}_\text{Parameter Learning}
\end{equation*}
Following \cite{Heckerman1995LearningData}, we can write structure learning as:
\begin{equation*}
    \underbrace{P(\mathcal{G} \, | \, \mathcal{D})}_\text{Posterior Distribution}
    \, \propto \,
    \underbrace{P(\mathcal{G})}_\text{Prior Distribution}
    \, \cdot \qquad
    \underbrace{P(\mathcal{D} \, | \, \mathcal{G}).}_\text{Likelihood}
\end{equation*}
The causal discovery problem \citep{Zanga2022RiskApproach, Zanga2022APractice,
Vowels2021DyaDiscovery} consists in recovering the true DAG $\mathcal{G}^*$ by
combining the available data $\mathcal{D}$ together with domain expert knowledge
$\mathcal{K}$. In this paper we tackle the causal discovery problem with a {\em
score-based approach}\footnote{Refer to \cite{Nandy2018High-dimensionalLearning}
for a more comprehensive overview of score-based structural learning.}, in which
a learning algorithm traverses the {\em search space}  $\mathbb{G}$ of DAGs
while looking for the DAG $\mathcal{G}^*$ maximizing the posterior
$P(\mathcal{G} \, | \, \mathcal{D})$. In the case of a uniform prior
distribution $P(\mathcal{G})$, the learning algorithm solves the following
optimization problem:
\begin{equation}\label{eq:causal_discovery}
    \mathcal{G}^* = \arg\max_{\mathcal{G} \in \mathbb{G}} P(\mathcal{G} \, | \, \mathcal{D})
    \qquad
    \text{with}
    \qquad
    P(\mathcal{G}) = \frac{1}{|\mathbb{G}|}.
\end{equation}
An example of such an algorithm is the \emph{Hill-Climbing}
\citep[HC;][]{chickering2002optimal} algorithm which starts from an empty DAG
$\mathcal{G}_0$ and evaluates iteratively the addition, deletion or reversal of
a candidate edge $\mathbf{e}_{XY}$ w.r.t. a {\em score function} $\mathcal{S}$,
often referred to as the \emph{scoring criterion}. At iteration $i$, the score
of the current candidate DAG $\mathcal{G}_{i}$ is compared to that of a new
candidate DAG $\mathcal{G}_{i+1}$. If the score $\mathcal{S}_{i+1}$ of
$\mathcal{G}_{i+1}$ is better than the score $\mathcal{S}_{i}$ of
$\mathcal{G}_{i}$, then the graph $\mathcal{G}_{i+1}$ becomes the new candidate
graph, otherwise $\mathcal{G}_{i}$ is retained as the best graph at iteration
$i+1$. This iterative process continues until the score of the current best DAG
cannot be increased further.

\subsection{Causal Discovery via Bagging}

The search space $\mathbb{G}$ of DAGs grows super-exponentially
\citep{Harary1973GraphicalEnumeration}: exhaustive search is unfeasible beyond
20--25 variables \citep{Koivisto2004ExactNetworks, rantanen2020discovering}.
Bagging (short for \emph{bootstrap aggregation}) provides a widely-used
alternative to full Bayesian model averaging that scales to larger numbers of
variables \citep{Friedman1999DataApproach, Friedman2003BeingNetworks}. In
essence, it is a two-step algorithm in which we:
\begin{enumerate}
  \item Bootstrap a set of DAGs $\bm{\mathcal{G}}$ learnt from a data set
    $\mathcal{D}$ and prior knowledge $\mathcal{K}$.
  \item Aggregate the bootstrapped DAGs $\bm{\mathcal{G}}$ to obtain an
    \emph{average} DAG $\smash{\mathcal{\hat{G}}}$.
\end{enumerate}
In the first step, we use non-parametric bootstrap
(\Cref{alg:non_parametric_boot}) to generate a collection of $n$ data sets
$\mathcal{D}_i$, with $i = 1, \ldots, n$, and to learn the associated DAGs
$\mathcal{G}_i$ while exploiting the prior knowledge $\mathcal{K}$. The DAGs
$\mathcal{G}_i$ form $\bm{\mathcal{G}}$  and can be learnt using any causal
discovery algorithm. In the second step, we use $\bm{\mathcal{G}}$ to estimate
the posterior expectation $\mathbb{E}[\mathcal{G} \; | \; \mathcal{D}]$. We
don't include $\mathcal{K}$ in the posterior expectation for brevity. There are
several examples of this type of approach in the literature.
\cite{Singh1997LearningData} combined the bootstrap step with the
Expectation-Maximization algorithm \citep{Lauritzen1995TheData} to learn BNs
from incomplete data. \cite{Friedman1999DataApproach} discussed a similar
approach and explored the difference between non-parametric and parametric
bootstrap. More recently, \cite{Rohekar2018BayesianBootstrap} proposed a new
recursive non-parametric bootstrap scheme, while
\cite{Sugahara2022BayesianSubbagging} investigated the efficacy of sub-bagging
from small samples.

\begin{algorithm2e}[H]
  \DontPrintSemicolon
  \SetKwProg{Fn}{Function}{}{}
  \caption{Non-Parametric Bootstrap.} \label{alg:non_parametric_boot}
  \KwIn{A data set $\mathcal{D}$, the prior knowledge $\mathcal{K}$ and the bootstrap iterations $n$.}
  \KwOut{The set of bootstrapped DAGs $\bm{\mathcal{G}}$.}
  \Fn{NonParametricBootstrap($\mathcal{D}, \mathcal{K}, n$)}{
      Initialize the bootstrapped DAG set $\bm{\mathcal{G}}$ to the empty set.\;
    \For{$i = 1, 2, \dots, n$}{
      Sample $|\mathcal{D}|$ observations with replacement from $\mathcal{D}$ to obtain $\mathcal{D}_i$.\;
      Learn a DAG $\mathcal{G}_i = \arg\max_{\mathcal{G} \in \mathbb{G}} P(\mathcal{G} \; | \; \mathcal{D}_i)$ given $\mathcal{D}_i$ and $\mathcal{K}$.\;
      Insert $\mathcal{G}_i$ into $\bm{\mathcal{G}}$.\;
    }
    \KwRet{$\bm{\mathcal{G}}$}\;
  }
\end{algorithm2e}

\subsection{The Impact of the Acyclicity Constraint}

As discussed in \cite{Scutari2013OnModelling}, the acyclicity constraint induces
a correlation between the edges present in any DAG learnt from data. This holds
for the uninformative uniform prior $P(\mathcal{G}) \propto 1$ and for any
posterior $P(\mathcal{G} \;|\; \mathcal{D})$, where the probabilistic
dependencies between the variables further increase the correlation between
edges that are part of causal pathways. These correlations are reflected in the
posterior expectation $\mathbb{E}[\mathcal{G} \; | \; \mathcal{D}]$ in general.
If cycles were allowed, the edges of $\mathcal{G}$, denoted as
$\mathbf{E}_{\mathcal{G}}$, would be independent of each other when the prior is
uniform. In particular, if we shift the focus from the posterior $P(\mathcal{G}
\;|\; \mathcal{D})$ of the graph $\mathcal{G}$ as a whole to the posterior
$P(\mathbf{E}_\mathcal{G} \;|\; \mathcal{D})$ of its edge set
$\mathbf{E}_\mathcal{G}$, we observe:
\begin{equation*}
  P(\mathcal{G} \; | \; \mathcal{D})
    = P(\mathbf{E}_\mathcal{G} \; | \; \mathcal{D})
    = \prod_{\mathclap{\mathbf{e}_{XY} \in \mathbf{E}_\mathcal{G}}}
        P(\mathbf{e}_{XY} \; | \; \mathbf{E}_\mathcal{G} \setminus \{ \mathbf{e}_{XY} \}, \mathcal{D})
    = \prod_{\mathclap{\mathbf{e}_{XY} \in \mathbf{E}_\mathcal{G}}}
        P(\mathbf{e}_{XY} \; | \; \mathcal{D})
\end{equation*}
However, since $\mathcal{G}$ is a DAG, the posterior of each edge
$P(\mathbf{e}_{XY} \; | \; \mathcal{D})$ is different from the posterior of each
edge \emph{given the others} $P(\mathbf{e}_{XY} \; | \; \mathbf{E}_\mathcal{G}
\setminus \{ \mathbf{e}_{XY} \}, \mathcal{D})$ in general. Moreover, including
$\mathbf{E}_\mathcal{G} \setminus \{ \mathbf{e}_{XY} \}$ in the probability term
makes the estimation of the marginal probability of each edge $\mathbf{e}_{XY}$
dependent on the distribution of all other possible edges, which, in turn, is
summarized by the expected value $\mathbb{E}[\mathcal{G} \; | \; \mathcal{D}]$.
Intuitively, we would prefer to approximate $P(\mathbf{e}_{XY} \; | \;
\mathcal{D})$ or $P(\mathbf{e}_{XY} \; | \; \mathbf{E}_\mathcal{G},
\mathcal{D})$ using a smaller subset $\mathbf{E}^*_\mathcal{G} \subset
\mathbf{E}_\mathcal{G}$ which contains the edges relevant for $\mathbf{e}_{XY}$.
This idea is inspired by \cite{Scutari2013OnModelling}, who proved that edges
that are not incident on a common vertex are uncorrelated for DAGs with up to 7
nodes in $P(\mathcal{G}) \propto 1$. He also showed by simulation that this
relationship holds for larger DAGs. These findings call for an extended
evaluation of potential subsets of $\mathbf{E}_\mathcal{G}$ containing edges
strongly correlated with $\mathbf{e}_{XY}$, which we discuss in the next
section.

\subsection{Threshold-based Model Averaging}

The current state-of-the-art averaging algorithm was initially described in
\cite{Singh1997LearningData} and independently proposed by
\cite{Friedman1999DataApproach, Imoto2002BootstrapRegression} with the addition
of a threshold. To our knowledge, no substantial modifications have been
investigated so far. Henceforth, we refer to this averaging algorithm as the
\emph{Threshold-based Model Averaging} (TMA) algorithm. This algorithm estimates
the probability $P(\mathbf{e}_{XY} \, | \, \mathcal{D})$ of including each edge
$\mathbf{e}_{XY}$ into the \emph{average DAG} $\smash{\mathcal{\hat{G}}}$ from
the set of bootstrapped DAGs $\bm{\mathcal{G}}$ and collects them into a matrix
$\mathbf{C}$ called the \emph{confidence matrix}. The aggregation step
constructs the DAG $\mathcal{\hat{G}}$ with an edge set
$\mathbf{E}_{\mathcal{\hat{G}}}$ such that it only contains the edges with the
highest confidence. As in \Cref{eq:causal_discovery}, $\bm{\mathcal{G}}$
approximates a sample from the posterior $P(\smash{\mathcal{\hat{G}}} \; | \;
\mathcal{D})$ and $\smash{\mathcal{\hat{G}}}$ is then an empirical estimate of
$\mathbb{E}[ \smash{\mathcal{G}} \;|\; \mathcal{D}]$. However, directly
aggregating all the edges in $\bm{\mathcal{G}}$ may result in a cyclic graph.
Therefore, we need to solve a constrained optimization problem to find the
$\mathcal{\hat{G}}$ with the maximal edge confidence set subject to the
acyclicity constraint. We start by redefining the constrained optimization
problem over the posterior $P(\mathbf{E}_{\mathcal{\hat{G}}} \; | \;
\mathcal{D})$:
\begin{equation}\label{eq:max_posterior_edges}
  \arg\max_{\mathcal{\hat{G}} \in \mathbb{G}}
    P(\mathcal{\hat{G}} \; | \; \mathcal{D})
  =
  \arg\max_{\mathcal{\hat{G}} \in \mathbb{G}}
    P(\mathbf{E}_\mathcal{\hat{G}} \; | \; \mathcal{D}).
\end{equation}
Indeed, \Cref{eq:max_posterior_edges} allows us to link the aggregation
algorithm over the edge set $\mathbf{E}_\mathcal{\hat{G}}$ and the posterior
maximization over the search space of DAGs $\mathbb{G}$. An approximation of the
optimization target $P(\mathbf{E}_\mathcal{\hat{G}} \; | \; \mathcal{D})$ is
given by the expected value of the posterior
$\mathbb{E}[\mathbf{E}_\mathcal{\hat{G}} \; | \; \mathcal{D}]$:
\begin{equation}\label{eq:def_expectation_edges}
  \arg\max_{\mathcal{\hat{G}} \in \mathbb{G}}
    P(\mathbf{E}_\mathcal{\hat{G}} \; | \; \mathcal{D})
  \approx
  \arg\max_{\mathcal{\hat{G}} \in \mathbb{G}}
    \mathbb{E}[\mathbf{E}_\mathcal{\hat{G}} \; | \; \mathcal{D}].
\end{equation}
The expectation $\mathbb{E}[\mathbf{E}_\mathcal{\hat{G}} \; | \; \mathcal{D}]$
plays a fundamental role in the aggregation step, providing a point estimate of
the posterior $P(\mathbf{E}_\mathcal{\hat{G}} \; | \; \mathcal{D})$ by
leveraging the information present in $\bm{\mathcal{G}}$. It is computed as the
frequency of inclusion of each edge $\mathbf{e}_{XY}$ in $\bm{\mathcal{G}}$:
\begin{equation}\label{eq:est_expectation_edges}
  \arg\max_{\mathcal{\hat{G}} \in \mathbb{G}}
    \mathbb{E}[\mathbf{E}_\mathcal{\hat{G}} \; | \; \mathcal{D}]
  =
  \arg\max_{\mathcal{\hat{G}} \in \mathbb{G}}
    \frac{1}{|\bm{\mathcal{G}}|}
    {\textstyle\sum_{\mathcal{G}}^{\bm{\mathcal{G}}}}
    {\textstyle\sum_{\mathbf{e}_{XY}}^{\mathbf{E}_\mathcal{G}}}
    \mathbbm{1}_{\mathcal{\hat{G}}}[\mathbf{e}_{XY} \; | \; \mathcal{D}],
\end{equation}
The outer summation ${\textstyle\sum_{\mathcal{G}}^{\bm{\mathcal{G}}}}$
aggregates the graphs $\mathcal{G}$ across the bootstrapped graphs
$\bm{\mathcal{G}}$, while the inner summation $
{\textstyle\sum_{\mathbf{e}_{XY}}^{\mathbf{E}_\mathcal{G}}}$ aggregates the
edges $\mathbf{e}_{XY}$ across the edge sets $\mathbf{E}_\mathcal{G}$. The
indicator function $\mathbbm{1}_{\mathcal{\hat{G}}}$ links the average DAG
$\mathcal{\hat{G}}$ to the objective function of the maximization problem.
Wrapping up \Crefrange{eq:max_posterior_edges}{eq:est_expectation_edges}, the
core of TMA is the following optimization problem:
\begin{equation}\label{eq:opt_tma}
  \arg\max_{\mathcal{\hat{G}} \in \mathbb{G}}
    P(\mathcal{\hat{G}} \; | \; \mathcal{D})
  \approx
  \arg\max_{\mathcal{\hat{G}} \in \mathbb{G}}
    \frac{1}{|\bm{\mathcal{G}}|}
    {\textstyle\sum_{\mathcal{G}}^{\bm{\mathcal{G}}}}
    {\textstyle\sum_{\mathbf{e}_{XY}}^{\mathbf{E}_\mathcal{G}}}
    \mathbbm{1}_{\mathcal{\hat{G}}}[\mathbf{e}_{XY} \; | \; \mathcal{D}].
\end{equation}
A hyperparameter $\alpha \in (0, 1)$ is used as a threshold for the values in
$\mathbf{C}$ to prevent irrelevant edges from being included into
$\mathcal{\hat{G}}$. \citet{Scutari2013IdentifyingNetworks} and
\citet{Liao2022OnNetworks} provide two data-driven choices for it value. Once
the threshold is applied, the construction of $\mathcal{\hat{G}}$ is performed
\emph{greedily}: edges are sorted by decreasing confidence and incrementally
added to $\mathcal{\hat{G}}$, skipping those inducing cycles (\Cref{alg:tma}).
Adapting TMA for cyclic models \citep{Nodelman2007ContinuousNetworks,
Bregoli2020Constraing-BasedNetworks,
Villa-Blanco2023Constraint-basedClassifiers} is straightforward. \\

\begin{algorithm2e}[H]
  \DontPrintSemicolon
  \SetKwProg{Fn}{Function}{}{}
  \SetKwComment{Comment}{}{}
  \caption{Threshold-based Model Averaging (TMA).} \label{alg:tma}
  \KwIn{A set of DAGs $\bm{\mathcal{G}}$ and a threshold $\alpha \in (0, 1)$.}
  \KwOut{An average DAG $\mathcal{\hat{G}}$.}
  \Fn{TMA($\bm{\mathcal{G}}, \alpha$)}{
    \Comment*[l]{\color{blue} \texttt{/* Phase I - Estimate confidence matrix. */}}
    $\mathbf{C} \gets \mathbf{0}$ \Comment*[r]{Initialize distribution parameters.}
    \For(\Comment*[f]{Iterate over DAGs.}){$\mathcal{G} \in \bm{\mathcal{G}}$}{
        $\mathbf{C}[X, Y] \gets \mathbf{C}[X, Y] + \mathbbm{1}_\mathcal{G}[X, Y], \quad \forall \, (\mathbf{e}_{XY}) \in \mathbf{E}_\mathcal{G}$ \Comment*[r]{Increment counts.}
    }
    $\mathbf{C} \gets \mathbf{C} \, / \, |\bm{\mathcal{G}}|$ \Comment*[r]{Normalize w.r.t.$\, |\bm{\mathcal{G}}|$.}
    \BlankLine

    \Comment*[l]{\color{blue} \texttt{/* Phase II - Threshold and sort by decreasing confidence. */}}
    $\mathbf{\tilde{\mathbf{C}}} \gets \big\{ (X, Y) \; | \; \mathbf{C}[X, Y] > \alpha \big\}$ \Comment*[r]{Set of $(X, Y)$ tuples s.t.$\, \mathbf{C}[X, Y] > \alpha$.}
    $\mathbf{\tilde{C}} \gets \argsort{\mathbf{\tilde{C}}}$ \Comment*[r]{Sort $(X, Y)$ tuples in $\mathbf{\tilde{\mathbf{C}}}$ by decreasing $\mathbf{C}[X, Y]$.}
    \BlankLine

    \Comment*[l]{\color{blue} \texttt{/* Phase III - Estimate average DAG. */}}
    $\mathcal{\hat{G}} \gets (\mathbf{V}, \varnothing)$ \Comment*[r]{Initialize an empty DAG.}
    \For(\Comment*[f]{Iterate over candidate edges.}){$(X, Y) \in \mathbf{\tilde{C}}$}{
        $\mathcal{G'} \gets \big(\mathbf{V}, \, \mathbf{E}_{\mathcal{\hat{G}}} \cup \{ \mathbf{e}_{XY} \} \big)$ \Comment*[r]{Compute new DAG.}
        \If(\Comment*[f]{Check if the new DAG is valid.}){{DAG}($\mathcal{G'}$)}{
            $\mathcal{\hat{G}} \gets \mathcal{G'}$ \Comment*[r]{Update current DAG.}
        }
    }
    \BlankLine

    \KwRet{$\mathcal{\hat{G}}$}
  }
\end{algorithm2e}

\section{Model Averaging on Higher-Order Interactions}\label{sec:model_averaging}

In this section, we propose a new averaging framework that leverages {\em
higher-order structures} to improve the TMA algorithm. We first introduce the
\emph{cover of a graph}.

\begin{definition}[Cover of a Graph]
  \label{def:cover}
  Let $\mathcal{G} = (\mathbf{V}, \mathbf{E})$ be a graph and $\mathcal{C} = \{
    \mathbf{E}_\varphi \, : \, \varphi \in \Phi \}$ be an indexed family of
  subsets $\mathbf{E}_\varphi \subset \mathbf{E}$ s.t. $\bigcup_{\varphi}^{\Phi}
  \mathbf{E}_\varphi = \mathbf{E}$, then $\mathcal{C}$ is said to be a cover of
  $\mathcal{G}$.
\end{definition}

Each index $\varphi$ corresponds to one and only one set $\mathbf{E}_\varphi$,
so we can use $\varphi$ for simplicity. In this context, $\Phi$ is an
\emph{index set} for a given graph $\Phi: \mathbb{G} \rightarrow \{ \varphi \}$.
While this definition $\varphi$ is flexible for identifying subsets of edges, it
does not allow to select the edges incident to a given vertex. We introduce the
notion of \emph{incident} cover for this purpose.

\begin{definition}[Incident Cover of a Graph]
  \label{def:incident_cover}
  Let $\mathcal{G} = (\mathbf{V}, \mathbf{E})$ be a graph and \linebreak
  \mbox{$\mathcal{C} = \{ \mathbf{E}_{\varphi_{_X}} \, : \, \varphi_{_X} \in
  \Phi, X \in \mathbf{V} \}$} be an indexed family of subsets
  $\mathbf{E}_{\varphi_{_X}} \subset \mathbf{E}$ s.t.
  \mbox{$\bigcup_{\varphi_{_X}}^{\Phi} \mathbf{E}_{\varphi_{_X}} = \mathbf{E}$}
  and every edge in $\mathbf{E}_{\varphi_{_X}}$ is incident on $X$, then
  $\mathcal{C}$ is said to be an incident cover of $\mathcal{G}$.
\end{definition}

Note that we only require $\varphi_{_X}$ to induce a subset
$\mathbf{E}_{\varphi_{_X}}$ of \emph{all} the edges incident to $X$, which
allows us to define $\Phi$ flexibly while retaining the \emph{locality} of each
$\varphi_{_X}$. Moreover, such a definition identifies one and only one
$\varphi_{_X}$ for each vertex $X \in \mathbf{V}$ for a given $\mathcal{G}$. We
refer to the index $\varphi_{_X}$ as the \emph{higher-order structure} of $X$
and, in turn, to the index set $\Phi_\mathcal{G}$ as the set of the higher-order
structures of $\mathcal{G}$.

\begin{example}[Parents-based Incident Cover]

  Let the incident cover $\mathcal{C}$ be indexed by \linebreak \mbox{$\Phi = \{
    (X, \Pi_X) \; | \; X \in \mathbf{V} \}$.} Each index is defined as
  $\varphi_{_X} = (X, \Pi_X)$ and each subset is \linebreak
  \mbox{$\mathbf{E}_{\varphi_{_X}} = \{ \mathbf{e}_{ZX} \; | \; Z \in \Pi_X
  \}$.} In other words, each set $\mathbf{E}_{\varphi_{_X}}$ in the cover
  $\mathcal{C}$ contains all and only the edges induced by the vertex $X$ and
  its parents $\Pi_X$. The key intuition is that $\varphi_{_X}$ identifies the
  relevant variables that interact with $X$, rather than just a single variable
  $Z$ as in the case of $\mathbf{e}_{ZX}$. \Cref{fig:higher_order} depicts this
  case for $\mathbf{e}_{BC}$ and $\mathbf{e}_{HI}$.

  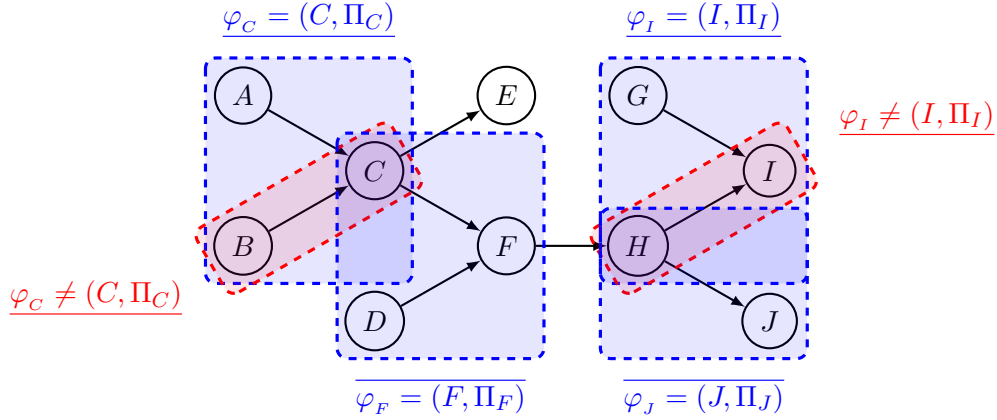
\begin{figure}[H]
    \centering
    \vspace{-1.25cm}
    \begin{tikzpicture}
      \node (0) at (+0.00, +1.00) {$A$};
      \node (1) at (+0.00, -1.00) {$B$};
      \node (2) at (+1.75, +0.00) {$C$};
      \node (3) at (+1.75, -2.00) {$D$};
      \node (4) at (+3.50, +1.00) {$E$};
      \node (5) at (+3.50, -1.00) {$F$};
      \node (6) at (+5.25, +1.00) {$G$};
      \node (7) at (+5.25, -1.00) {$H$};
      \node (8) at (+7.00, +0.00) {$I$};
      \node (9) at (+7.00, -2.00) {$J$};

      \draw[->] (0) to (2);
      \draw[->] (1) to (2);
      \draw[->] (2) to (4);
      \draw[->] (2) to (5);
      \draw[->] (3) to (5);
      \draw[->] (5) to (7);
      \draw[->] (6) to (8);
      \draw[->] (7) to (8);
      \draw[->] (7) to (9);

      \draw[
        rounded corners, dashed, very thick,
        blue, fill=blue, fill opacity=0.10
      ]
      (-0.50, +1.50) rectangle (+2.25, -1.50)
      node[draw=none, blue, fill opacity=1, pos=0.5, yshift=+2.00cm]
      {$\underline{\varphi_{_C} = (C, \Pi_C)}$};

      \draw[
        rotate around={120:(0.875,-0.50)},
        rounded corners, dashed, very thick,
        red, fill=red, fill opacity=0.10
      ]
      (+0.375, +1.00) rectangle (+1.375, -2.00)
      node[draw=none, red, fill opacity=1, pos=-0.85, yshift=2.00cm]
      {$\underline{\varphi_{_C} \neq (C, \Pi_C)}$};

      \draw[
        rounded corners, dashed, very thick,
        blue, fill=blue, fill opacity=0.10
      ]
      (+1.25, +0.50) rectangle (+4.00, -2.50)
      node[draw=none, blue, fill opacity=1, pos=0.5, yshift=-2.00cm]
      {$\overline{\varphi_{_F} = (F, \Pi_F)}$};

      \draw[
        rounded corners, dashed, very thick,
        blue, fill=blue, fill opacity=0.10
      ]
      (+4.75, +1.50) rectangle (+7.50, -1.50)
      node[draw=none, blue, fill opacity=1, pos=0.5, yshift=+2.00cm]
      {$\underline{\varphi_{_I} = (I, \Pi_I)}$};

      \draw[
        rotate around={120:(6.125,-0.50)},
        rounded corners, dashed, very thick,
        red, fill=red, fill opacity=0.10
      ]
      (+5.625, +1.00) rectangle (+6.625, -2.00)
      node[draw=none, red, fill opacity=1, pos=+1.85, yshift=-2.00cm]
      {$\underline{\varphi_{_I} \neq (I, \Pi_I)}$};

      \draw[
        rounded corners, dashed, very thick,
        blue, fill=blue, fill opacity=0.10
      ]
      (+4.75, -0.50) rectangle (+7.50, -2.50)
      node[draw=none, blue, fill opacity=1, pos=0.5, yshift=-1.50cm]
      {$\overline{\varphi_{_J} = (J, \Pi_J)}$};
  \end{tikzpicture}
  \vspace{-1.25cm}
  \caption{Higher-order interactions are highlighted in blue, compared to
    lower-order ones in red. Note that $\varphi_{_J}$ is a special case where
    higher- and lower-order interactions coincide.}
  \label{fig:higher_order}
\end{figure}
\end{example}

The causal discovery problem is formulated as an optimization problem with an
objective function depending on $P(\mathcal{G} \; | \; \mathcal{D})$. We now
redefine such an objective function following $\Phi_\mathcal{G}$:
\begin{equation}
  \label{eq:max_posterior_hoi}
  \mathcal{G}^*
    = \arg\max_{\mathcal{G} \in \mathbb{G}} P(\mathcal{G} \; | \; \mathcal{D})
    = \arg\max_{\mathcal{G} \in \mathbb{G}} P(\mathbf{E}_\mathcal{G} \; | \; \mathcal{D})
    = \arg\max_{\mathcal{G} \in \mathbb{G}} P(\Phi_\mathcal{G} \; | \; \mathcal{D})
\end{equation}
As for the TMA algorithm, we rely on the expected value
$\mathbb{E}[\Phi_\mathcal{\hat{G}} \; | \; \mathcal{D}]$ over the set of
bootstrapped DAGs $\bm{\mathcal{G}}$ to provide an estimate of
$P(\Phi_\mathcal{\hat{G}} \; | \; \mathcal{D})$, following the bagging approach.

\begin{definition} [Causal Discovery by Higher-Order Interactions]
  \label{def:causal_hoi}
  Let $\mathcal{G}^*$ be the true but unknown DAG, $\mathbb{G}$ the space of
  DAGs and $\mathcal{D}$ the data set induced by $\mathcal{G}^*$. The causal
  discovery problem on higher-order interactions consists in solving:
  \begin{equation}\label{eq:opt_gma}
    \mathcal{G}^* =
      \arg\max_{\mathcal{G} \in \mathbb{G}}
      {\textstyle\sum_{\varphi_{_X}}^{\Phi_\mathcal{G}}}
      P(\varphi_{_X} \; | \; \mathcal{D})
  \end{equation}
  where the summation ${\textstyle\sum_{\varphi_{_X}}^{\Phi_\mathcal{G}}}$
  aggregates higher-order structures $\varphi_{_X}$ across $\Phi_\mathcal{G}$.
\end{definition}

We leverage \Cref{def:incident_cover} to explain \Cref{eq:opt_gma}, providing an
estimate of the global posterior $P(\Phi_\mathcal{\hat{G}} \; | \; \mathcal{D})$
in terms of each local posterior $P(\varphi_{_X} \; | \; \mathcal{D})$. We start
by approximating the posterior with an expectation over the distribution of the
higher-order structures:
\begin{equation}
  \arg\max_{\mathcal{\hat{G}} \in \mathbb{G}}
    P(\Phi_\mathcal{\hat{G}} \; | \; \mathcal{D})
  \approx
  \arg\max_{\mathcal{\hat{G}} \in \mathbb{G}}
    \mathbb{E}
    [\Phi_\mathcal{\hat{G}} \; | \; \mathcal{D}].
\end{equation}
This step is an adaptation of \Cref{eq:def_expectation_edges} in the context of
the higher-order structures $\Phi_\mathcal{\hat{G}}$. We estimate
$\mathbb{E}[\Phi_\mathcal{\hat{G}} \; | \; \mathcal{D}]$ by counting the
occurrences of each $\varphi_{_X}$ across $\bm{\mathcal{G}}$:
\begin{equation}\label{eq:est_expectation_hoi}
  \arg\max_{\mathcal{\hat{G}} \in \mathbb{G}}
   \mathbb{E}
   [\Phi_\mathcal{\hat{G}} \; | \; \mathcal{D}]
  =
  \arg\max_{\mathcal{\hat{G}} \in \mathbb{G}}
    \frac{1}{|\bm{\mathcal{G}}|}
    {\textstyle\sum_{\mathcal{G}}^{\bm{\mathcal{G}}}}
    {\textstyle\sum_{\varphi_{_X}}^{\Phi_\mathcal{G}}}
    \mathbbm{1}_\mathcal{\hat{G}}
    [\varphi_{_X} \; | \; \mathcal{D}]
\end{equation}
Finally, we rearrange the normalization term $1 / |\bm{\mathcal{G}}|$ and the
outer summation ${\textstyle\sum_{\mathcal{G}}^{\bm{\mathcal{G}}}}$ inside the
inner summation ${\textstyle\sum_{\varphi_{_X}}^{\Phi_\mathcal{G}}}$ to obtain
the local posterior $P(\varphi_{_X} \; | \; \mathcal{D})$ for each $\varphi_{_X}
\in \Phi_\mathcal{G}$:
\begin{equation}\label{eq:hoi_estimate}
  \arg\max_{\mathcal{G} \in \mathbb{G}}
    \frac{1}{|\bm{\mathcal{G}}|}
    {\textstyle\sum_{\mathcal{G}}^{\bm{\mathcal{G}}}}
    {\textstyle\sum_{\varphi_{_X}}^{\Phi_\mathcal{G}}}
    \mathbbm{1}_\mathcal{\hat{G}}
    [\varphi_{_X} \; | \; \mathcal{D}]
  =
  \arg\max_{\mathcal{\hat{G}} \in \mathbb{G}}
    {\textstyle\sum_{\varphi_{_X}}^{\Phi_\mathcal{G}}}
    P(\varphi_{_X} \; | \; \mathcal{D})
\end{equation}
From \Cref{def:causal_hoi} and \Cref{eq:hoi_estimate}, we derive the
\emph{Generalized Model Averaging} (GMA) algorithm. Pseudocode is given in
\Cref{alg:gma}. It is a generic template for causal discovery on higher-order
interactions based on $\bm{\mathcal{G}}$ and the index set $\Phi$:
\begin{enumerate}[align=left,label={Phase \Roman{*} - },leftmargin=3\parindent]
  \item Estimating each $P(\varphi_{_X} \; | \; \mathcal{D})$ following
    \Cref{def:causal_hoi},
  \item Sorting each $\varphi_{_X}$ depending on the estimated $P(\varphi_{_X}
    \; | \; \mathcal{D})$ terms,
  \item Maximizing the posterior probability $P(\Phi_\mathcal{\hat{G}} \; | \;
    \mathcal{D})$ greedily s.t. $\mathcal{\hat{G}}$ is acyclic.
\end{enumerate}

\begin{algorithm2e}[H]
  \DontPrintSemicolon
  \SetKwProg{Fn}{Function}{}{}
  \SetKwComment{Comment}{}{}
  \caption{Generalized Model Averaging (GMA).}
  \label{alg:gma}
  \KwIn{A set of DAGs $\bm{\mathcal{G}}$, an index set $\Phi$ and a threshold $\alpha \in (0, 1)$.}
  \KwOut{An average DAG $\mathcal{\hat{G}}$.}
  \Fn{GMA($\bm{\mathcal{G}}, \Phi, \alpha$)}{
    \Comment*[l]{\color{blue} \texttt{/* Phase I - Estimate posterior distribution. */}}
    $\bm{\Phi} \gets \mathbf{0}$ \Comment*[r]{Initialize distribution parameters.}
    \For(\Comment*[f]{Iterate over DAGs.}){$\mathcal{G} \in \bm{\mathcal{G}}$}{
      $\bm{\Phi}[\varphi_{_X}] \gets \bm{\Phi}[\varphi_{_X}] + \mathbbm{1}_\mathcal{G}[\varphi_{_X}], \quad X \in \mathbf{V}, \varphi_{_X} \in \Phi_\mathcal{G}$ \Comment*[r]{Increment counts.}
    }
    $\bm{\Phi} \gets \bm{\Phi} \; / \; \sum_{\varphi_{_X}} \bm{\Phi}[\varphi_{_X}]$ \Comment*[r]{Estimate $P(\varphi_{_X} \; | \; \mathcal{D})$.}
    \BlankLine

    \Comment*[l]{\color{blue} \texttt{/* Phase II - Threshold and sort by decreasing posterior. */}}
    $\bm{\tilde{\Phi}} \gets \big\{ \varphi_{_X} \; | \; \bm{\Phi}[\varphi_{_X}] > \alpha \big\}$ \Comment*[r]{Set of $\varphi_{_X}$ s.t.$\, \bm{\Phi}[\varphi_{_X}] > \alpha$.}
    $\bm{\tilde{\Phi}} \gets \argsort{\bm{\tilde{\Phi}}}$ \Comment*[r]{Sort $\varphi_{_X}$ in $\bm{\tilde{\Phi}}$ by decreasing $\bm{\Phi}[\varphi_{_X}]$.}
    \BlankLine

    \Comment*[l]{\color{blue} \texttt{/* Phase III - Estimate average DAG. */}}
    $\mathcal{\hat{G}} \gets (\mathbf{V}, \varnothing)$ \Comment*[r]{Initialize an empty DAG.}
    \For(\Comment*[f]{Iterate over higher-order structures.}){$\varphi_{_X} \in \bm{\tilde{\Phi}}$}{
      $\mathcal{G'} \gets \big(\mathbf{V}, \, \mathbf{E}_{\mathcal{\hat{G}}} \cup \mathbf{E}_{\varphi_{_X}} \big)$ \Comment*[r]{Compute new DAG.}
      \If(\Comment*[f]{Check if new DAG is valid.}){DAG($\mathcal{G'}$)}{
          $\mathcal{\hat{G}} \gets \mathcal{G'}$ \Comment*[r]{Update current DAG.}
      }
    }
    \BlankLine

    \KwRet{$\mathcal{\hat{G}}$}
  }
\end{algorithm2e}

\newpage

\noindent The time and space complexity of GMA are as follows. Let $k =
|\bm{\mathcal{G}}|$, $n = |\mathbf{V}|$ and $m = |\mathbf{E}|$:

\begin{itemize}
  \item The space complexity of $\bm{\mathcal{G}}$ is $\mathcal{O}(kn^2)$. The
    worst-case scenario corresponds to the case where the DAGs are dense.
    However, causal discovery algorithms usually assume that the true DAG is
    sparse, in which case the space complexity is $\Theta(km)$ where $m$ is the
    average number of edges over $\bm{\mathcal{G}}$.
  \item The space complexity of the $\bm{\Phi}$ is $\mathcal{O}(kn)$. In
    general, the complexity of $\bm{\Phi}$ is $\mathcal{O}(n \cdot 2^n)$ since
    $\varphi_{_X}$ takes values in the power set $2^\mathbf{V}$. However, we
    store values only for the observed $\varphi_{_X}$. Hence, the worst case is
    where we observe exactly $k$ distinct $\varphi_{_X}$ for each vertex $X$.
    Therefore, the final space complexity is $\mathcal{O}(kn)$. Furthermore, we
    are guaranteed that $\bm{\Phi}$ is sparse since $k \ll 2^{n}$ when the time
    complexity of obtaining $\bm{\mathcal{G}}$ is super-exponential, which is
    the case for most real-world applications.
\end{itemize}
In summary, the worst-case space complexity of \Cref{alg:gma} is
$\mathcal{O}(kn^2)$.

Assuming $\bm{\Phi}$ is a sparse matrix with $\mathcal{O}(n)$ access time, the
time complexities for the most relevant lines are as follows:
\begin{itemize}
  \item Line 3-4, the \texttt{for} cycle requires $\mathcal{O}(kn^2)$ to
    accumulate the counts.
  \item Line 5, the element-wise sparse multiplication takes
    $\mathcal{O}(kn^2)$.
  \item Line 7, sorting $\bm{\hat{\Phi}}$ is equivalent to sort a set with
    cardinality $kn$, hence $\mathcal{O}(kn \cdot \log (kn))$.
  \item Line 8-12, the \texttt{for} cycle iterates $\mathcal{O}(kn)$ times and
    the inner $DAG$ algorithm takes $\mathcal{O}(n + m)$, with $m$ representing
    the number of edges of $\mathcal{\hat{G}}$. Therefore, the overall time
    complexity of this block in the worst-case scenario is $\mathcal{O}(kn \cdot
    (n + m))$.
\end{itemize}
In conclusion, the worst-case time complexity of \Cref{alg:gma} is
$\mathcal{O}(kn \cdot (n + m))$.

\section{Experiments} \label{sec:experiments}

We compared the existing TMA algorithm w.r.t. specific higher-order structures.
In particular, we investigated the impact of the choice of higher-order
structure by evaluating three GMA variants that differ only in the definition of
the index set $\Phi$: the set of \emph{parents} (PMA), the set of
\emph{children} (CMA) and the set of \emph{incident vertices} (IMA). These
methods are evaluated against the DAGs of the reference BN models reported in
\Cref{tab:ref_models}, often used as a benchmark for simulation studies. To
guarantee complete transparency and reproducibility, experimental results are
obtained using open-source software made publicly available
\href{https://github.com/AlessioZanga/higher-order-interactions}{here}.

\subsection{Experimental Design}

For each reference model $\mathcal{B}$, we sample multiple data sets
$\mathcal{D}$ using forward sampling. Each data set has sample size
$|\mathcal{D}| = |\Theta|\rho$, where $|\Theta|$ is the number of parameters of
$\mathcal{B}$ and $\rho$ a sample ratio coefficient in $[0.1, 0.2, 0.5, 1.0,
2.0, 5.0]$. For each sample ratio, we generate a training set and a test set for
in-sample and out-of-sample validation. Following
\Cref{alg:non_parametric_boot}, each training set is then resampled to produce a
bootstrap set $\bm{\mathcal{G}}$ of size 100 using HC and the BIC scoring
criterion for causal discovery. Finally, we aggregated the bootstrap set
$\bm{\mathcal{G}}$ into the average DAG $\mathcal{\hat{G}}$ following the
different model averaging algorithms discussed earlier. The threshold $\alpha$
is tuned with a grid search and set to 0.5 for TMA and to $1 / (|\mathbf{V}| -
1)$ for PMA, CMA and IMA \citep{Liao2022OnNetworks}.

We computed the following metrics with respect to the true DAG underlying
$\mathcal{B}$:
\begin{itemize}
  \item Bayesian Scoring Criterion (BIC): \cite{schwarz1978estimating}
    introduced the BIC as an estimate of the complexity of the model. It is part
    of the \emph{penalized log-likelihood} family, where the penalization term
    depends on both the number of  parameters of the given model and the sample
    size of the data set from which the model has been learnt:
    \begin{equation*}
      \mathit{BIC}(\Theta \; | \; \mathcal{G}, \mathcal{D}) =
          \log \mathcal{L}(\Theta \; | \; \mathcal{G}, \mathcal{D})
          - \frac{1}{2} |\Theta| \log{|\mathcal{D}|}.
    \end{equation*}
    In the following, \texttt{scaled\_in\_bic} and \texttt{scaled\_out\_bic} are
    reported as in-sample and out-of-sample estimates of BIC, respectively. To
    allow for comparison between different sample ratios, a min-max scaler is
    applied intra-sample ratio. Note that BIC does not require the true causal
    graph to be given.

  \item Structural Hamming Distance (SHD): The cardinality of the symmetric
    difference, denoted as $\Delta$, between the edge sets of $\mathcal{G}$ and
    $\mathcal{H}$ \citep{tsamardinos2006max}:
    \begin{equation*}
      \mathit{SHD}(\mathcal{G}, \mathcal{H}) =
        \lvert\mathbf{E}_\mathcal{G} \; \Delta \; \mathbf{E}_\mathcal{H}\rvert =
        \lvert
          (\mathbf{E}_\mathcal{G} \setminus \mathbf{E}_\mathcal{H}) \cup
          (\mathbf{E}_\mathcal{H} \setminus \mathbf{E}_\mathcal{G})
        \rvert.
    \end{equation*}
  \item Balanced Accuracy \& F1 Score: Differences in the adjacency matrices of
    two graphs can be represented using a \emph{confusion matrix}, especially
    when $\mathcal{G}$ is the \emph{predicted} one and $\mathcal{H}$ is the
    \emph{true} one, deriving true positives (TP), true negatives (TN), false
    positives (FP) and false negatives (FN) with the same semantics used in
    classification problems. To take into account sparsity, we rely on Balanced
    Accuracy (BA) and the F1 Score:
    \begin{align*}
      &\mathit{BA} =
        \frac{1}{2} \bigg( \frac{\mathit{TP}}{\mathit{TP} + \mathit{FN}} +
        \frac{\mathit{TN}}{\mathit{TN} + \mathit{FP}} \bigg),&
      &\mathit{F1} = \frac{2\mathit{TP}}{2\mathit{TP} + \mathit{FP} + \mathit{FN}}.
    \end{align*}
\end{itemize}

\begin{table}[H]
  \begin{center}
  \begin{tabular}{l|cccccc}
  \toprule
  \textbf{Model} &
  \textbf{Parameters} &
  \textbf{Vertices} &
  \textbf{Edges} &
  \textbf{Avg. M.B.} &
  \textbf{Avg. Deg.}
  \\ \midrule
  \texttt{ALARM}      & 509    & 37   & 46   & 3.51 & 2.49 \\
  \texttt{WIN95PTS}   & 574    & 76   & 112  & 5.92 & 2.95 \\
  \texttt{INSURANCE}  & 1008   & 27   & 52   & 5.19 & 3.85 \\
  \texttt{ANDES}      & 1157   & 223  & 338  & 5.61 & 3.03 \\
  \texttt{HAILFINDER} & 2656   & 56   & 66   & 3.54 & 2.36 \\
  \texttt{PIGS}       & 5618   & 441  & 592  & 3.66 & 2.68 \\
  \bottomrule
  \end{tabular}
  \caption{Summary statistics of the DAGs of the reference BNs sorted by
    increasing number of parameters. Average Markov blanket (Avg. M.B.) and
    average degree (Avg. Deg.) are computed as the mean of the \emph{Markov
    blanket} \citep{koller2009probabilistic} and adjacent vertices across each
    vertex in the vertex set. }
  \label{tab:ref_models}
  \end{center}
\end{table}

\begin{figure}[H]
  \includegraphics[width=\linewidth]{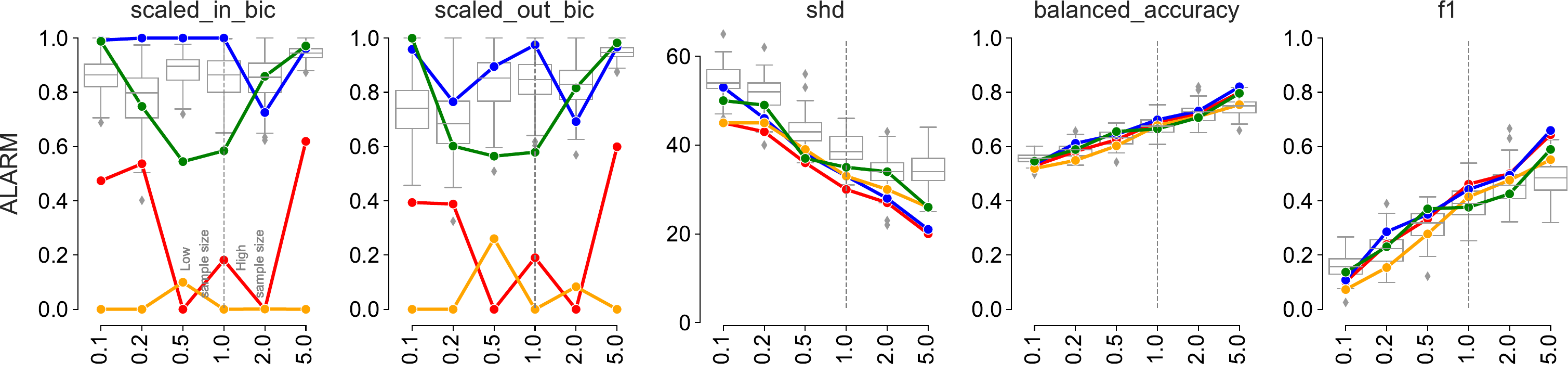}
  \includegraphics[width=\linewidth]{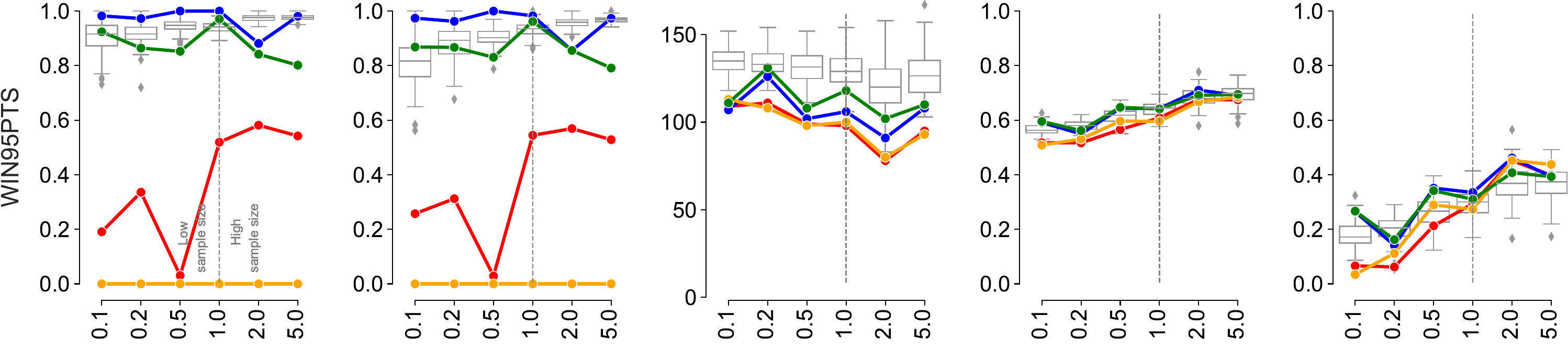}
  \includegraphics[width=\linewidth]{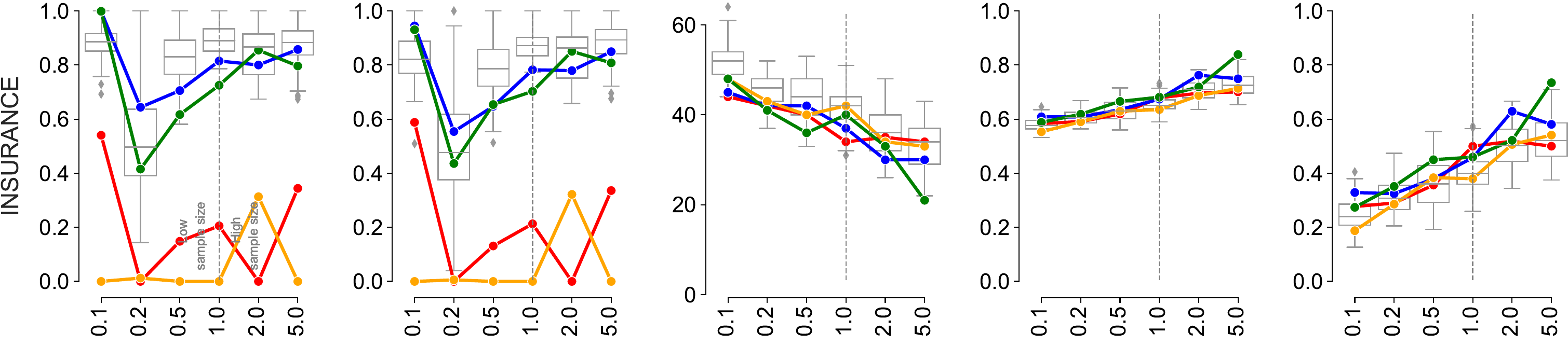}
  \includegraphics[width=\linewidth]{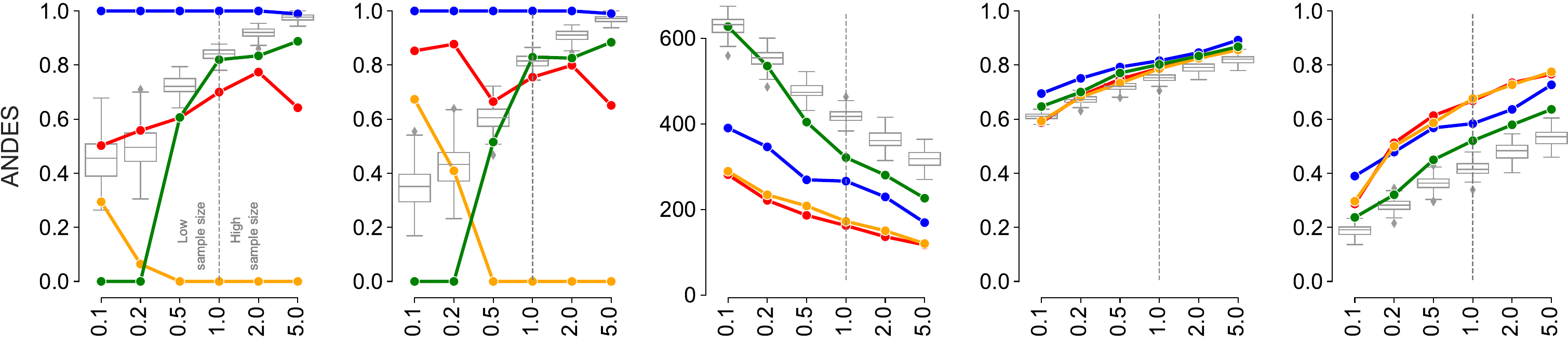}
  \includegraphics[width=\linewidth]{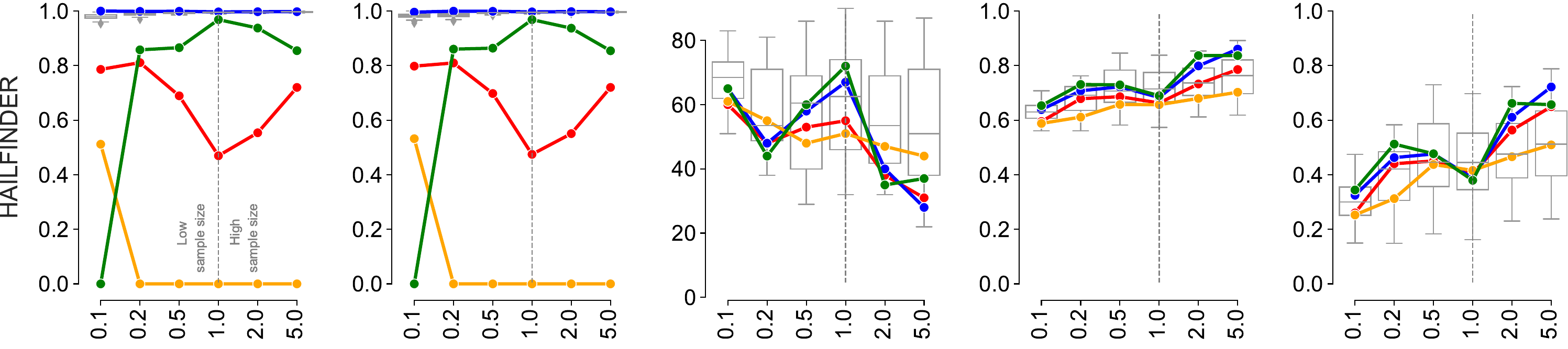}
  \includegraphics[width=\linewidth]{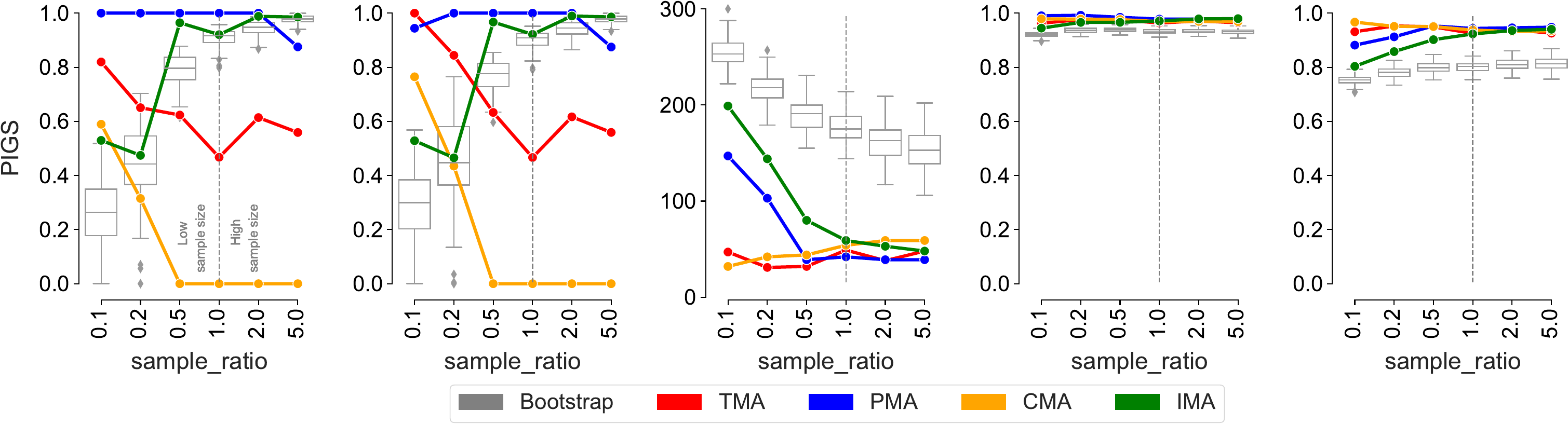}
  \vspace{-0.80cm}
  \caption{Bagging for the reference BNs in \Cref{tab:ref_models}.}
  \label{fig:wo-score-main}
\end{figure}

\subsection{Experimental Results}

The experimental results are shown in \Cref{fig:wo-score-main}. Each subfigure
comprises five panels corresponding to the metrics described in the previous
subsection. On the horizontal axis, we reported the sample ratio in increasing
order; on the vertical axis, we can find the scale of the given metric. SHD
takes values in [0, $+\infty$), where the lower, the better; other metrics take
values in [0, 1], where higher values imply better performance. The proposed
algorithms (PMA in blue, CMA in yellow and IMA in green) are compared against
the TMA algorithm (in red), which is taken as a baseline. The values of the
metrics for the individual bootstrapped DAGs $\bm{\mathcal{G}}$ are also shown
in grey using box-plots for comparison. The vertical gray-dashed line separates
low-sample-size (left) and large-sample-size (right) results.

The PMA algorithm achieves the best in-sample and out-of-sample BIC across every
reference DAG, although it is sometimes tied with IMA. This is particularly
apparent in low sample size regimes and reference models with many variables. In
some cases, for instance, in \texttt{ALARM} and \texttt{WIN95PTS}, it performs
better than any other approach, including the bootstrap set itself. In some
cases, we observe higher SHD values for PMA and IMA. Other balanced metrics are
in line with the baseline and the bootstrap set, suggesting that the structural
differences are inflated by the unbalanced proportion between extra and missing
edges. Both TMA and CMA achieve good structural performance but fail to match
PMA in terms of score, recovering less-likely DAGs compared to the other
alternatives. Finally, IMA recovers acceptable average DAGs in the case of small
models and high sample sizes, but its metrics degrade rapidly with the reference
model's complexity.

\section{Discussion \& Conclusions}\label{sec:conclusions}

In this work, we investigated learning an averaged DAG via bagging in the
context of causal discovery. In particular, we leveraged higher-order structures
to propose a novel theoretical framework, which we used to design a new
aggregation algorithm for combining bootstrapped DAGs. We performed a
large-scale simulation study on a varied set of reference DAGs to assess the
performance of the proposed solution: it dominates state-of-the-art alternatives
in structural accuracy and goodness-of-fit, especially under high-dimensional
settings and in low-sample size regimes. However, there is still room for
improvement. Firstly, considering the topological orderings of the bootstrapped
DAGs could mitigate the structural differences in the observed higher-order
structures. Secondly, weighting higher-order structures by the scoring criterion
could provide additional insights into their goodness of fit. Finally,
alternative data-driven thresholds could result in better denoising performance.

\acks{Alessio Zanga is funded by F. Hoffmann-La Roche Ltd.}


\end{document}